\documentclass[runningheads]{llncs}
\usepackage{graphicx}
\usepackage{enumerate}
\usepackage{enumitem}
\usepackage{xcolor}
\usepackage{multirow}
\usepackage{subfigure}
\usepackage{booktabs}
\usepackage{hyperref}
\usepackage[misc]{ifsym}

\usepackage{amsmath}
\usepackage{amsfonts}

\newcommand{\zhanyu}[1]{}
\newcommand{\guanjie}[1]{}
\newcommand{\chumeng}[1]{}
\newcommand{\nop}[1]{}

\begin{document}
\title{FDTI: Fine-grained Deep Traffic Inference with Roadnet-enriched Graph}
\toctitle{FDTI: Fine-grained Deep Traffic Inference with Roadnet-enriched Graph}
\author{Zhanyu~Liu\inst{1} \and
Chumeng~Liang\inst{1} \and
Guanjie~Zheng\inst{1}(\Letter) \and
Hua~Wei\inst{2}
}
\tocauthor{Zhanyu~Liu, Chumeng~Liang, Guanjie~Zheng, Hua~Wei}
\authorrunning{Z. Liu et al.}
\institute{
Shanghai Jiao Tong University, Shanghai, China
\email{\{zhyliu00,gjzheng\}@sjtu.edu.cn,caradryan2022@gmail.com}\\
\and
Arizona State University, Tempe, USA.\\
\email{hua.wei@asu.edu}
}
\maketitle              

\begin{abstract}
This paper proposes the fine-grained traffic prediction task (e.g. interval between data points is 1 minute), which is essential to traffic-related downstream applications. 
Under this setting, traffic flow is highly influenced by traffic signals and the correlation between traffic nodes is dynamic. 
 As a result, the traffic data is non-smooth between nodes, and hard to utilize previous methods which focus on smooth traffic data.
 To address this problem, we propose \underline{\textbf{F}}ine-grained \underline{\textbf{D}}eep \underline{\textbf{T}}raffic \underline{\textbf{I}}nference, termed as \textbf{FDTI}. 
 Specifically, we construct a fine-grained traffic graph based on traffic signals to model the inter-road relations.
 Then, a physically-interpretable dynamic mobility convolution module is proposed to capture vehicle moving dynamics controlled by the traffic signals. 
 Furthermore, traffic flow conservation is introduced to accurately infer future volume.
 Extensive experiments demonstrate that our method achieves state-of-the-art performance and learned traffic dynamics with good properties. To the best of our knowledge, we are the first to conduct the city-level fine-grained traffic prediction.


\end{abstract}
\section{Introduction}
Traffic prediction is an important part of an intelligent traffic system and benefits downstream tasks.
Some downstream tasks are sensitive to the granularity of prediction results, such as traffic signal control, congestion discovery, and route planning.
Taking traffic signal control as an example, predictions on the 1-minute level could timely evaluate the impact of the incoming traffic signal and improve traffic policy because the interval of traffic signal change is approximately 1 minute~\cite{koonce2008traffic}.
Previous deep methods~\cite{song2020spatial,diao2019dynamic,cirstea2022towards} focus on the coarse-grained traffic data.
However, it remains unexplored that utilizes deep methods to solve traffic prediction tasks under the fine-grained setting.

\begin{figure}
    \centering
    \includegraphics[width=1\linewidth]{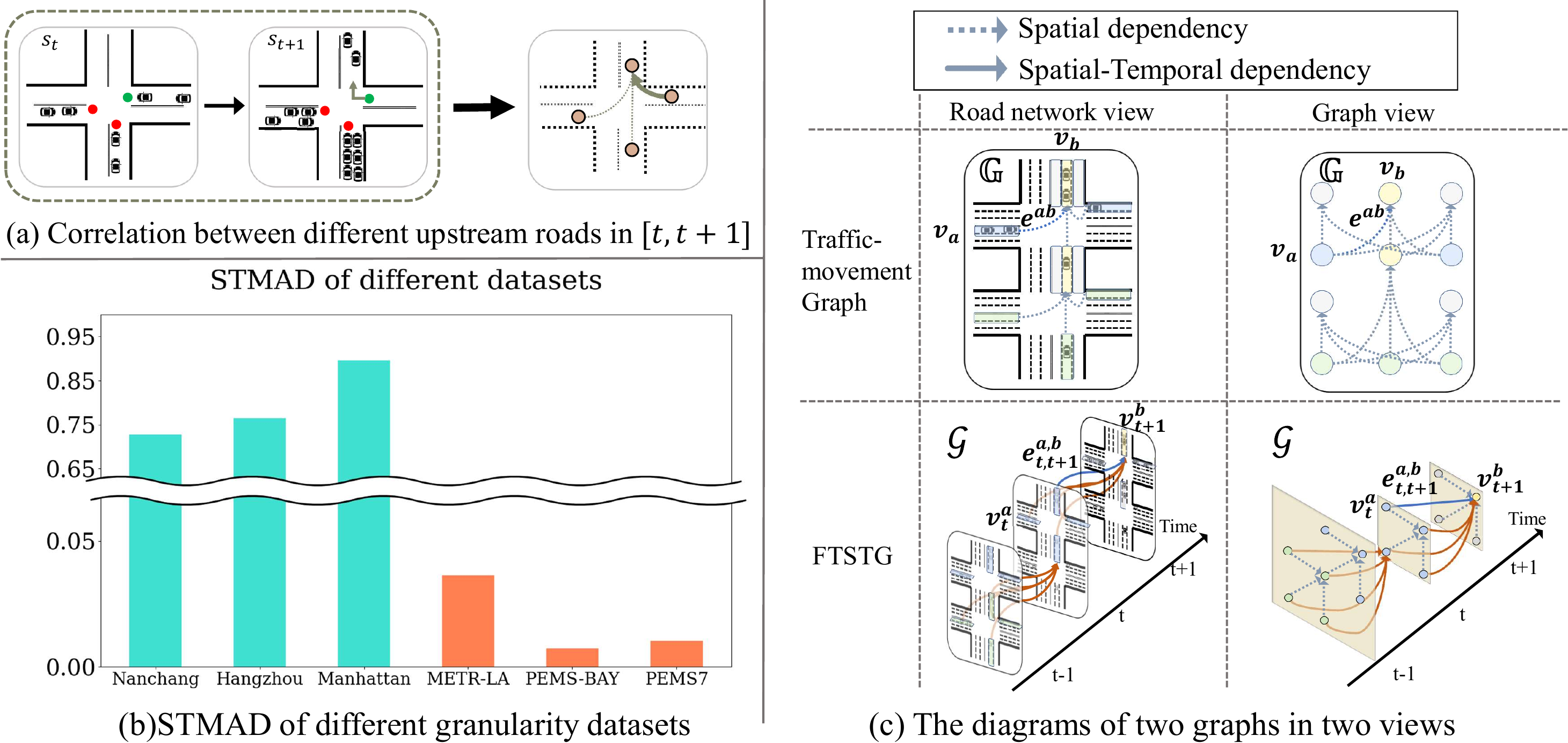}
    \caption{(a) The traffic signal determines the traffic flow, thereby determining the correlation between roads. (b) Our fine-grained data is much more unsmooth than previous datasets. (Low STMAD indicates smoother and the wavy line indicates the omitted space). (c) Diagrams of Traffic-movement graph and FTSTG.}
    \label{fig:intro}
\end{figure}

Under the fine-grained setting, the traffic flow is determined by traffic signals~\cite{lei2022modeling}.
When the signal turns green, the vehicles could flow into downstream roads. 
As a result, the correlations between these roads are strong under the traffic prediction context, which is shown in Fig~\ref{fig:intro}(a). 
However, previous research ignores the explicit highly dynamic correlation between nodes under the fine-grained settings and utilizes static graphs~\cite{li2017diffusion,yu2017spatio} or data-driven graphs~\cite{bai2020adaptive,shao2022decoupled,rao2022fogs} to aggregate the knowledge of nodes.

Due to the highly dynamic correlations resulting from the traffic signals, spatial neighbors do not have similar traffic volumes.
Therefore, as shown in Fig~\ref{fig:intro}(b), the fine-grained traffic data is non-smooth, which is evaluated by Spatial Temporal Mean Average Distance (STMAD) defined in this paper.
Previous methods have satisfying results on coarse-grained smooth datasets.
However, since smoothing is the essential nature of the GCN design~\cite{chen2020measuring}, experiments show previous methods still make smooth predictions on the nonsmooth fine-grained data which leads to big errors.

To better model the dynamic correlations and tackle the non-smoothness of the data under the fine-grained setting, we propose a model called \underline{\textbf{F}}ine-grained \underline{\textbf{D}}eep \underline{\textbf{T}}raffic \underline{\textbf{I}}nference (\textbf{FDTI}). First, to adapt the characteristic that traffic signal controls traffic flow in fine-grained traffic inference, we construct a \textbf{Fine-grained Traffic Spatial-Temporal Graph (FTSTG)}. Specifically, we build a road network feature enriched multi-layer traffic graph, in which each layer represents a time frame as shown in Fig~\ref{fig:intro}(c). Edges inside the graph represent traffic flow links between two nodes at adjacent time frames, which are controlled by traffic signals. 
Then, we propose a \textbf{Dynamic Mobility Convolution Network} to induce consistency with the traffic policy on FTSTG. People can make a metaphor between a traffic network and a water flow network, in which the traffic signal is similar to the tap controlling the flow. 
Based on the previous two modules, we further infer the traffic volume of each node following \textbf{Flow Conservative Traffic State Transition}.
Our contribution can be summarized as follows.
\begin{itemize}[leftmargin=*]
    \item To the best of our knowledge, we are the first to complete the city-level fine-grained traffic prediction, which is important in intelligent traffic systems and will enable efficient and in-time traffic policy-making and other downstream tasks.
    \item We propose a model named Fine-grained Deep Traffic Inference (FDTI) to incorporate the dynamic spatial temporal dependency caused by traffic signals and then the future traffic is inferred in a flow-conservative perspective. 
    \item Extensive experiments on traffic datasets have shown the superior performance of our proposed method.
    Graph smoothness analysis is conducted based on our proposed metric STMAD, which explains the mechanism of how other baselines fail under the fine granularity setting.
    
\end{itemize}

\section{Related Work}
\textbf{Conventional Traffic Prediction.} 
Traffic prediction research draws lots of attention~\cite{xie2020urban}, while conventional methods focus on statistical methods.
Kalman filter based methods~\cite{okutani1984dynamic,lippi2013short} show good results for short-term traffic volume prediction.
ML methods such as SVM~\cite{nikravesh2016mobile} built on non-linear relationships achieve better performance.
The spatial dependency is modeled by methods such as Bayesian Network~\cite{zhu2016short}, and probabilistic model~\cite{akagi2018fast}.
However, they have not exploited the rich spatial information enough.

\noindent
\textbf{Deep Spatial-Temporal Traffic Prediction.} 
The utilization of graph convolutional networks (GCNs)~\cite{kipf2016semi,velivckovic2017graph} contributes significantly to the advancement of spatial-temporal traffic prediction.
DCRNN~\cite{li2017diffusion}, STGCN~\cite{yu2017spatio}, GSTNet~\cite{fang2019gstnet}, STDN~\cite{yao2019revisiting}, STFGNN~\cite{li2021spatial}, LSGCN~\cite{huang2020lsgcn} combines modules such as diffusion, GCN and GRU to model the spatial and temporal relations.
Recently many adaptive methods for spatial-temporal data have been proposed. 
Methodologies such as Graph Wavenet~\cite{wu2019graph}, AGCRN~\cite{bai2020adaptive}, GMAN~\cite{zheng2020gman},  FC-GAGA~\cite{oreshkin2021fc}, D2STGNN~\cite{shao2022decoupled}, HGCN~\cite{guo2021hierarchical}, ST-WA~\cite{cirstea2022towards}, DSTAGNN~\cite{lan2022dstagnn} utilize techniques such as node embedding and attention to reconstruct the adaptive adjacent matrix and fuse the temporal long term relation.
MDTP~\cite{fang2021mdtp}, MTGNN~\cite{wu2020connecting}, and DMSTGCN~\cite{han2021dynamic} utilize multimodal data to help forecast the traffic.
Z-GCNET~\cite{chen2021z} introduces time-aware persistent homology.
STGODE~\cite{fang2021spatial}, STG-NCDE~\cite{choi2022graph}, STDEN~\cite{ji2022stden} use differential equations to model the traffic.
However, most of those methods would utilize enormous parameters on learning graphs with node embeddings which ignores the influence of traffic signals between different nodes.
\cite{qu2022forecasting}, \cite{ouyang2020fine} researches the fine-grained volume inference. 
However, they focus on spatial-fine-grained grid-based data and utilize CNN-based methods, which can not be applied to our temporal-fine-grained graph-based data. A recent work~\cite{lei2022modeling} focuses on fine-grained graph-based traffic prediction, which incorporates a similar state transition function but uses a different setting of missing data.
\section{Preliminaries}

\begin{definition}[Traffic-Movements Grap] 
We model the traffic system as a traffic-movement graph $\mathbb{G} = (\mathbb{V},\mathbb{E})$ where $\mathbb{V}$ is the set of $N$ traffic-movements~\cite{zheng2019learning} 
and $\mathbb{E}$ is the set of connections between traffic movements. 
Each traffic-movement $v^i$ is a set of lanes with the same moving direction $d^i \in \{Left, Straight, Right\}$.
Each directed edge $e^{ij} $ denotes the link from traffic movement $v^i$ to traffic movement $v^j$. Fig~\ref{fig:intro}(c) shows a sample traffic-movement graph $\mathbb{G}$ generated from the real traffic system. 
\end{definition}

\begin{definition}[Traffic State] 
The traffic state $\textbf{x}_t^i$ of a traffic-movement $v^i$ at timestamp $t$ includes various measurements such as speed and volume. Thus, the traffic state of the whole system is represented as $\textbf{X}_t=\{\textbf{x}_t^1,\textbf{x}_t^2,\cdots,\textbf{x}_t^N\}$.
In this paper, we mainly focus on traffic volume, defined as the number of vehicles on the traffic-movement $v^i$ at timestamp $t$.
The time granularity of the traffic volume is 1 minute.
\end{definition}

\begin{definition}[Roadnet-enriched Feature] 
Roadnet is an abbreviation for road network.
The roadnet-enriched feature indicates the road-network-related feature that helps infer future traffic states.
It contains the traffic signal (described as green signal time $p_i$) and the static information of traffic-movements (e.g., length $l^i$ and direction $d^i$).
Foramally, the system-wise roadnet-enriched feature is represented as $\textbf{S}_t=\{\textbf{s}_t^1,\textbf{s}_t^2,\cdots,\textbf{s}_t^N\}$ where $\textbf{s}_t^i=\{p_t^i,l^i,d^i\}$ is the features of traffic-movement $v^i$ at time $t$.
\end{definition}

\subsection{Problem definition}

\begin{problem}[One-step inference]
Given a city-level traffic system $\mathbb{G} = (\mathbb{V},\mathbb{E})$, the goal is to learn a model $f$ to perform traffic inference of next time step $\textbf{X}_{t+1}$ based on traffic state observations $\textbf{X}_{t-T+1:t}$ and roadnet-enriched feature $\textbf{S}_{t-T+1:t}$ of previous $T$ time steps. 
Formally, the problem is defined as
\begin{equation}
\hat{\textbf{X}}_{t+1}=f(\textbf{X}_{t-T+1:t},\textbf{S}_{t-T+1:t}).
\end{equation}
\end{problem}

\begin{problem}[Q-step inference]
Based on one-step state inference, Q-step state inference can be achieved by performing one-step inference Q times. Formally, this problem could be denoted as
\begin{equation}
\hat{\textbf{X}}_{t'+1}=f(\hat{\textbf{X}}_{t'-T+1:t'},\textbf{S}_{t'-T+1:t'}), t' = t+1, \cdots, t+Q
\end{equation}

\end{problem}

Here $\hat{\textbf{X}}_{t'-T+1:t'}$ is the input of function $f$ and it could include both predicted value and ground truth value.

\section{Method}
\begin{figure*}[!thb]
    \centering
    \includegraphics[width=1\linewidth]{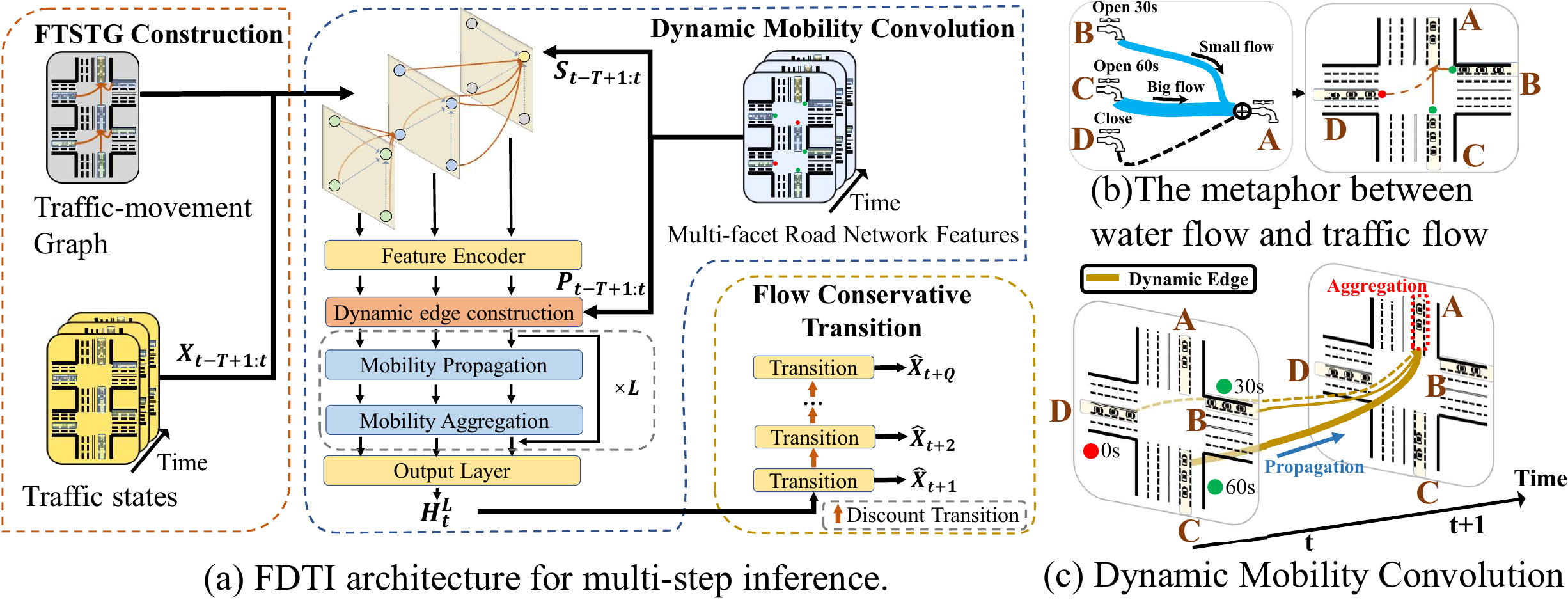}
    \caption{Diagrams of Fine-grained Deep Traffic Inference (FDTI).}
    \label{fig:simple_diagram}
\end{figure*}

To solve the defined problem, we propose \underline{\textbf{F}}ine-grained \underline{\textbf{D}}eep \underline{\textbf{T}}raffic \underline{\textbf{I}}nference (\textbf{FDTI}) as illustrated in Fig~\ref{fig:simple_diagram}(a). 
Firstly, traffic states and roadnet-enriched features are organized to construct FTSTG, which represents the traffic node in a graph with multiple time layers.
Then, a dynamic mobility convolution is conducted to model the traffic flow transition via dynamic edges.
Lastly, the model predicted the traffic flow, and the future traffic volume of each node is inferred on considering the conservation of the traffic system.

\subsection{Fine-grained Traffic Spatial-Temporal Graph}

\subsubsection{Graph Construction}

In this section, we introduce the construction of Fine-grained Traffic Spatial-Temporal Graph as shown in Fig~\ref{fig:intro}(c). 
For the sake of understanding, we make an analogy between a traffic flow network controlled by traffic signals and a water network controlled by taps (as shown in Fig~\ref{fig:simple_diagram}(b)).
For the simple water network,  the water volume of A (denoted as $x_t^A$ at step $t$) can be inferred based on the flow-in volume $\iota_t^A$ and flow-out volume $o_t^A$ as 
\begin{equation}
x_{t+1}^A = x_{t}^A + \iota_{t}^A - o_{t}^A.
\label{eq:inout_water}
\end{equation}
By considering the spatial dependency, $\iota_{t}^A$ and $o_{t}^A$ can be calculated with the data $B$,$C$, and $D$, as shown below. 
\begin{equation}
    \iota_{t}^A = \sigma(x_{t}^B, \tau_{t}^B, x_{t}^C, \tau_{t}^C, x_{t}^D, \tau_{t}^D)
    \label{eq:in_water}
\end{equation}
\begin{equation}
    o_{t}^A = \phi(x_{t}^A,\tau_{t}^A)
    \label{eq:out_water}
\end{equation}
where $\sigma$ and $\phi$ calculate the flow-in volume and flow-out volume based on $x$ and the turn-on time of the water tap $\tau$.

A traffic system can be represented similarly. Traffic signals can naturally substitute the role of taps in the water network. 
Eqs.~\eqref{eq:inout_water},\eqref{eq:in_water} and \eqref{eq:out_water} show that the states of spatial neighbors of A at timestamp $t$
($x_t^B,x_t^C,x_t^D$) are highly related to the state of A at timestamp $t+1$ ($x_{t+1}^A$), which inspires us how to construct Fine-grained Traffic Spatial-Temporal Graph (FTSTG). Formally, FTSTG is denoted as $\mathcal{G} = \{\mathcal{V},\mathcal{E}\}$, where each vertex $v_t^i \in \mathcal{V}$ denotes the node $i$ at timestamp $t$. 
Here the size of $\mathcal{V}$ is calculated as $|\mathcal{V}|=N\times T$ where $N$ is the number of traffic movements and $T$ is the total number of timestamps. We model the spatial-temporal dependency by the edges.
\begin{equation}
    <v_t^i,v_{t+1}^j>=\left\{
    \begin{array}{rcl}
    1& &<i,j>\in \mathbb{E}\ or\ i=j\\
    0& &otherwise
    \end{array} \right.
\end{equation}
where $\mathbb{E}$ is the edge set of the graph $\mathbb{G}$. (1) We add the edge between spatial neighbors of different time layers. 
(2) We add edges between the same node of the adjacent time layer.
(3) There is no edge inside the same time layer, which is the key difference between FTSTG and STSGCN~\cite{song2020spatial}.

The roadnet-enriched features $\textbf{S}_t=\{P_t,l,d\}$ along with the historical traffic states $\textbf{X}_t$ serve as the input of each node.
There are two reasons why the features $\textbf{S}_t$ help forecast future traffic.
Firstly, The green signal time $P_t$ controls the traffic flow according to Eqs.~\eqref{eq:in_water} and \eqref{eq:out_water} and thus significantly influence the future traffic volume.
Secondly, the length $l$ and the turning direction $d$ influence the volume distribution of the traffic node since longer roads tend to have more traffic volume, and right-turning lanes tend to have less traffic volume.

\subsection{Dynamic Mobility Convolution}

To capture the spatial temporal dependencies, we propose Dynamic Mobility Convolution on FTSTG, which is shown in Fig~\ref{fig:simple_diagram}(c). This builds a model that approximates the function $\sigma$ and $\phi$ in Eqs.~\eqref{eq:in_water} and \eqref{eq:out_water}.

\subsubsection{Dynamic Edge Construction}

To utilize the spatial temporal dependency, a traditional methodology is to apply graph convolution operation on the FTSTG. However, as shown in Fig~\ref{fig:simple_diagram}(b), the traffic flow between different nodes is highly related to the green signal time. 
Inspired by this fact, we add Dynamic Edge Construction on FTSTG as shown in Fig~\ref{fig:simple_diagram}(c). The dynamic edges are related to the green signal time of each vertex and could represent the traffic flow mobility. A higher weight of dynamic edges indicates higher mobility of traffic flow. Thus, by denoting the edge weight of $<v_t^i,v_{t'}^j>$ as $w_{t,t'}^{i,j}$, and the green signal time of $v_t^i$ as $p_t^i$, we build the edge of FTSTG as follow.
\begin{equation}
    w_{t,t'}^{i,j}=\left\{
    \begin{array}{rcl}
    \frac{p_t^i}{t'-t} & &if\ <v_t^i,v_{t'}^j>\ \in \mathcal{E}\\
    0& &otherwise
    \end{array} \right.
\end{equation}

\subsubsection{Mobility Propogation and Mobility Aggregation}

After the Dynamic Edge Construction, we conduct Mobility Propagation and Mobility Aggregation based on the idea of GraphSAGE\cite{hamilton2017inductive}, which is a representative inductive graph learning method. The key idea of Mobility Propagation and Mobility Aggregation is that the hidden states of FTSTG represent the traffic flow and the dynamic edge represents the traffic flow mobility. Then one propagation-aggregation operation layer is simulating the process that vehicles flow into downstream nodes once, which is also an inductive operation. The output of the $l$-th layer can be derived as follows.

\begin{equation}
    H_{t,i}^l = Agg(H_{t,i}^{l-1},Prop( \{H_{t-1,j}^{l-1}|j\}))
\end{equation}
Here $j\in\{k|<v_{t-1}^k,v_t^i>\in\mathcal{E}\}$, which are the spatial neighbors of $i$ and  $i$ itself at previous adjacent timestamp. For Mobility Propagation, we take the dynamic edge into the operation. Formally we can write.
\begin{equation}
    \hat{H}_{t,i}^l=  f(\{H_{t-1,j}^{l-1}\cdot w_{t-1,t}^{j,i}|j\}).
\end{equation}

Then Mobility Aggregation is conducted to aggregate the result of Mobility Propagation and hidden states, which could be formulated as.
\begin{equation}
    H_{t,i}^l=  g(H_{t,i}^{l-1},\hat{H}_{t,i}^l).
\end{equation}

For $f$ and $g$, multiple functions such as \emph{MEAN($\cdot$), POOL($\cdot$), Concat($\cdot$), FC($\cdot$)} could be chosen. Furthermore, we add residual links~\cite{he2016deep} between adjacent blocks

A key observation is that one layer of propagation and aggregation feeds all the required input contained in Eqs.~\eqref{eq:inout_water},\eqref{eq:in_water}, and \eqref{eq:out_water} to state $x_{t+1}^i$. 
This means the number of layers of propagation and aggregation is equal to the number of historical horizons that are aggregated to $H_{t+1,i}^l$.
Typically, the model only needs to consider several adjacent horizons and get good results, which keeps consistent with the fact that only traffic states of adjacent time stamps are useful in the fine-grained traffic inference scenario.

\subsection{Flow Conservative Traffic State Transition}

\subsubsection{Traffic Flow Prediction}

The Dynamic Mobility Convolution learns representations $\textbf{H}^L_t$ that capture the fine-grained spatial temporal dynamics.
Based on that, we can predict the flow features, i.e., the out number $\hat{\textbf{O}}_t$ and in number $\hat{\textbf{I}}_t$ by using two fully connected layers.

\begin{equation}
    \hat{\textbf{O}}_t=FC(\textbf{H}^L_t), \hat{\textbf{I}}_t=FC(\textbf{H}^L_t)
\end{equation}

\subsubsection{One-step Traffic Inference}

After the out number $\hat{\textbf{O}}_t$ and in number $\hat{\textbf{I}}_t$ is predicted, the future traffic could be inferred in a flow conservative perspective.
Eq.~\eqref{eq:onestep} shows the transition from current observation $\textbf{X}_t$ to the inference of next timestamp $\hat{\textbf{X}}_{t+1}$ based on the out number $\hat{\textbf{O}}_t$ and in number $\hat{\textbf{I}}_t$.
\begin{equation}
    \hat{\textbf{X}}_{t+1} = \textbf{X}_t + \hat{\textbf{I}}_t - \hat{\textbf{O}}_t
    \label{eq:onestep}
\end{equation}
This shows a flow-conservative perspective for traffic inference.
Intuitively, the volume of a node would stay conserved if there are no vehicles driving in or driving out.
Hence, by considering each node as a closed traffic system, we only need to focus on the number of the drive-in and drive-out vehicles for future volume inference.
This is a key difference between FDTI and other conventional approaches to traffic prediction. 
Conventional approaches focus on capturing the numerical pattern based on mechanisms such as convolution and ignore the conservative traffic state transition which is the intrinsic dynamics.

\subsubsection{Multi-step Traffic Inference}

For multi-step traffic volume inference, the future multi-faceted features $S_{t+1:t+P}$ is predefined since the traffic signal policy is set in advance. Thus, we can simply apply traffic state transition Eq.~\eqref{eq:onestep} multiple times. However, multi-step inference still suffers from error accumulation~\cite{bengio2015scheduled} when a vertex takes inaccurate information from the previous one. Thus, we propose a discounting mechanism to reduce the accumulated error. The discounted multi-step traffic volume inference could be formulated as Eq.~\eqref{eq:multilayer} where $\lambda$ denotes the discounting factor. 
\begin{equation}
    \hat{\textbf{X}}_{t+Q} = \textbf{X}_t + \sum_{q=0}^{Q-1}\lambda^q(\hat{\textbf{I}}_{t+q} - \hat{\textbf{O}}_{t+q})
    \label{eq:multilayer}
\end{equation}

We choose MSE loss as the objective function for the one-step flow feature inference to train the model. Thus the loss function of FDTI for flow prediction can be formulated as
\begin{equation}
\mathcal{L} = \frac{1}{NT}\sum_{i=1}^N\sum_{t=1}^T(\iota_t^i-\hat{\iota}_t^i)^2 + \frac{1}{NT}\sum_{i=1}^N\sum_{t=1}^T(o_t^i-\hat{o}_t^i)^2.
\end{equation}

\section{Experiment}

\subsection{Experiment Settings}
\begin{figure}[!tb]
    \centering 
    \resizebox{0.8\linewidth}{!}{
    \begin{tabular}{ccc}
        \includegraphics[width=0.33\linewidth]{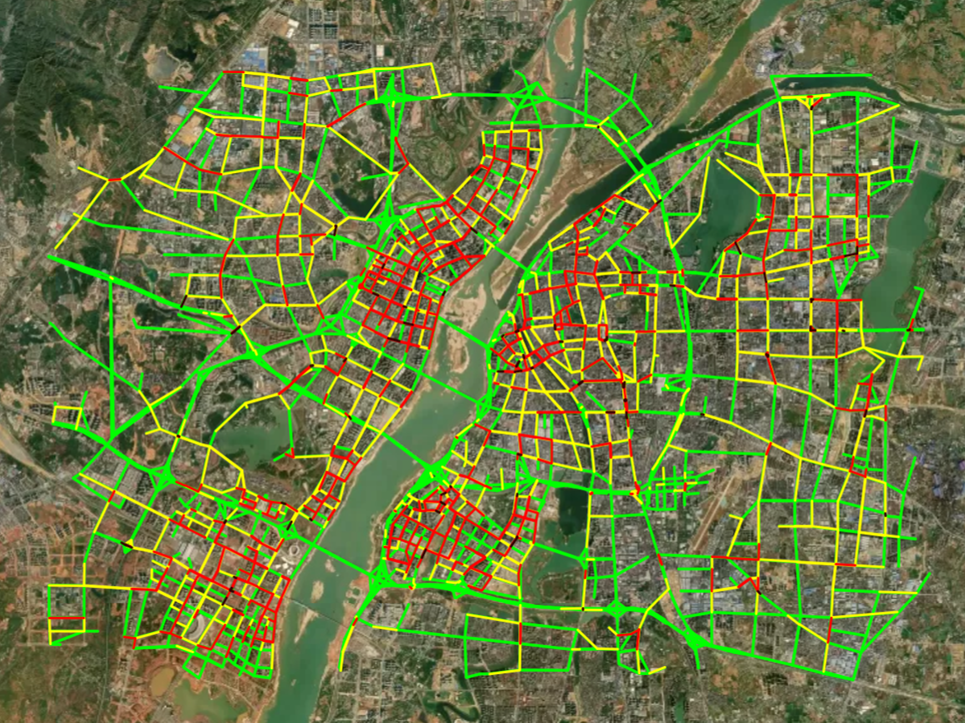}
        &  \includegraphics[width=0.33\linewidth]{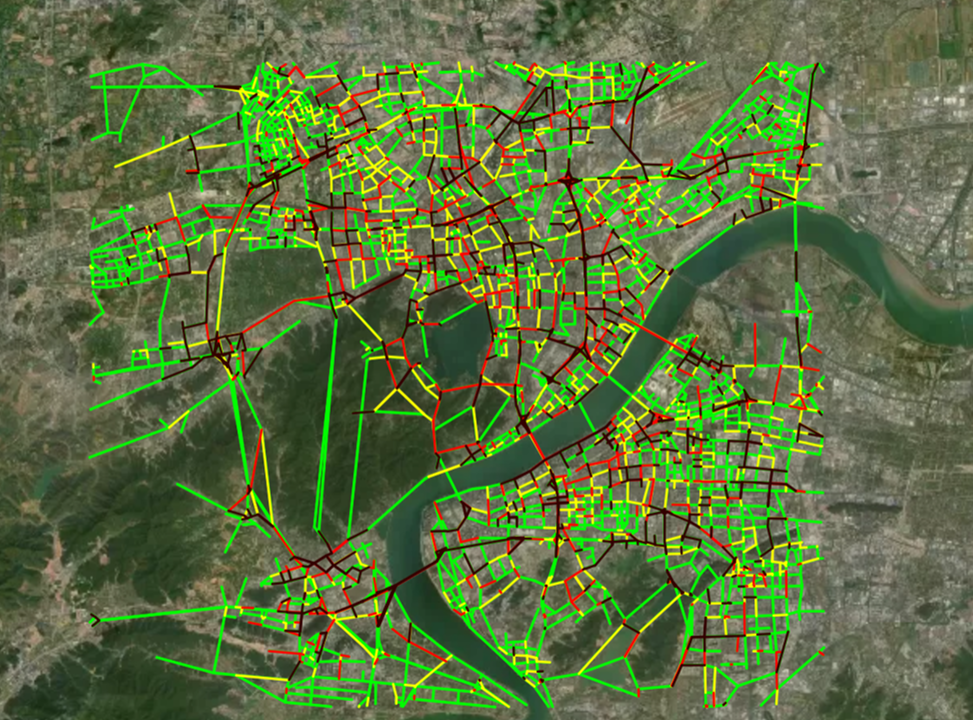}
        &  \includegraphics[width=0.33\linewidth]{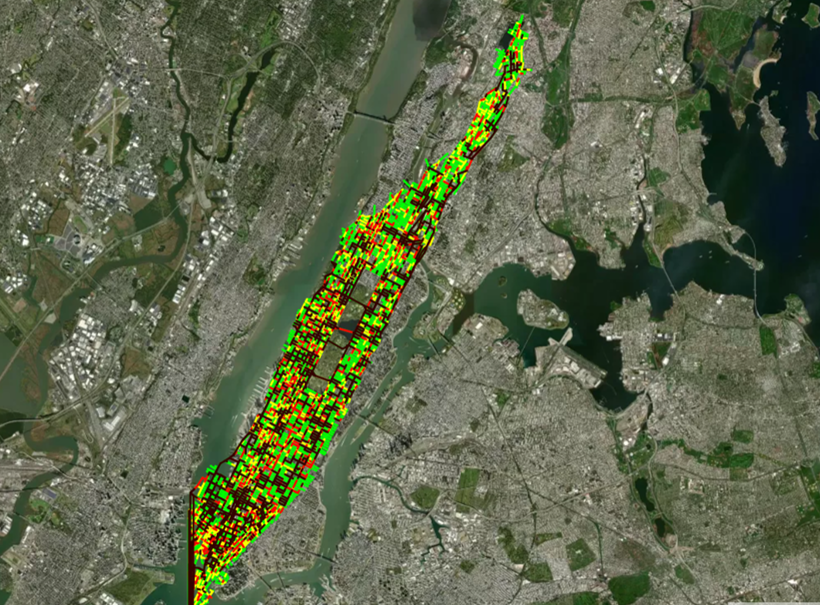}\\
        (a) Nanchang & (b)Hangzhou & (c) Manhattan
    \end{tabular}
    }
    \caption{The running screenshots of the traffic simulator. }
    \label{fig:simulator}
\end{figure}
\subsubsection{Datasets}
\begin{table}[!tb]
\centering
\caption{Details of datasets}
\label{tab:datasets}
\resizebox{0.6\linewidth}{!}{
\begin{tabular}{c|c|c|c}
\toprule
 City & Intersections & Nodes & Connections between nodes\\
\midrule
Nanchang & 2,048 & 18,072&73,170 \\
\midrule
Hangzhou & 3,819 & 32,772&76,788\\
\midrule
Manhattan &3,938 & 40,026&107,442\\
\midrule
Hangzhou-Small & 211& 1,944 & 4,770 \\
\bottomrule
\end{tabular}
}
\end{table}
We evaluate our model on three city-wide large-scale datasets of Nanchang, Manhattan, Hangzhou, and one small dataset Hangzhou-Small.
Current city traffic data is sparse, coarse-grained, and lacks traffic signal information.
Hence, utilizing real roadnet data and vehicle trajectory data as input, we collect 1-hour fine-grained data from the wildly-used traffic simulator of KDDCUP2021~\cite{liang2022cblab}. 
The roadnet data of these three cities is extracted from OpenStreetMap\footnote{https://www.openstreetmap.org/}.
The vehicle trajectory of Manhattan is processed real data from~\cite{wei2019survey}.
The vehicle trajectories of Hangzhou and Nanchang are from the real information reported by the traffic police. 
The details of the four datasets are shown in Table~\ref{tab:datasets}. 
The running traffic screenshots of Nanchang, Hangzhou, and Manhattan in the traffic simulator are shown in Fig~\ref{fig:simulator}.
Code and data are released in~\url{https://github.com/zhyliu00/FDTI}.

\subsubsection{Setup of Experiments}

\begin{itemize}[leftmargin=*]
    \item \emph{Data preprocessing}: The first 10 minutes are used to initialize the road network with sufficient vehicles. The remaining part of the data is split by the ratio of 6:2:2 in chronological order for training, validation, and testing.
    \item \emph{Network Structure}: 
    In mobility propagation, we use max-pooling as the function $f$.
    For the function $g$ in mobility aggregation, we concatenate $H_{t,i}^{l-1}$ and $\hat{H}_{t,i}^{l}$ and feed them into a fully-connected layer. 
    $tanh$ is used as the activation function.
    we set the hidden dimension of graph convolution as 256 and the graph convolution layers $L$ as 4.
    \item \emph{Training \& Evaluating}: The model is optimized by Adam optimizer for at most 500 epochs. The learning rate is set to 0.0005. We evaluate the performance of related models by RMSE and MAPE.
    $$
    RMSE = \sqrt{\frac{1}{s}\sum_{i=1}^s(y_i-\hat{y}_i)^2},\ \  MAPE = \frac{1}{s}\sum_{i=1}^s |\frac{y_i-\hat{y}_i}{{y_i}}|
    $$
\end{itemize}

\subsubsection{Compared Methods}

We compare FDTI with the following baselines. For the sake of fairness, all of the baselines except HA and ARIMA take the same node feature ([$v_t^i$, $p_t^i$, $l^i$, one hot coding for $d^i$]) and all of them are fine-tuned. Four types of baselines are compared.
\begin{itemize}[leftmargin=*]
    \item \textit{Traditional methods}: 
    \textbf{HA} is historical average method, and \textbf{ARIMA}~\cite{williams2003modeling} is a statistical time series analysis method.
    
    \item \textit{Basic Machine Learning models}: 
    \textbf{LSTM}~\cite{hochreiter1997long} is a classic RNN-based model for series analysis.
    \textbf{LR} exploits the linear correlations between data. 
    \textbf{XGBoost}~\cite{chen2016xgboost} is a competitive method based on boosting-tree.
    
    \item \textit{Convolution-Kernel-based STGNN}: 
    \textbf{DCRNN}~\cite{li2017diffusion} and \textbf{STGCN}~\cite{yu2017spatio} use GCN, GRU, and diffusion techniques to model the spatial and temporal dependencies.
    \textbf{STDEN}~\cite{ji2022stden} is a physics-based ODE method that models the traffic flow.
    
    \item \textit{Adaptive-based STGNN}: 
    \textbf{AGCRN}~\cite{bai2020adaptive}, \textbf{DGCRN}~\cite{li2021dynamic}, \textbf{D2STGNN}~\cite{shao2022decoupled} focuses on learning the dynamic graph by various methods such as node embeddings and learnable traffic pattern matrix. 
    \textbf{FOGS}~\cite{rao2022fogs} utilize node2vec-based methods to learn the graph. 
    However, these methods could not run on three large-scale datasets due to the out-of-memory error.
    They are only evaluated on the HangzhouSmall dataset.
\end{itemize}

\subsection{Overall Performance}

\begin{table*}[!tb]
\caption{Performance comparison between FDTI and baselines on three large datasets. All adaptive methods can not run on these large-scale datasets due to the huge memory consumption. The lower the RMSE and MAPE are, the better. Horizon means the number of forecasting steps and one horizon means one minute. FDTI achieves the best performance}
\label{tab:performance}
\resizebox{0.99\linewidth}{!}{
\begin{tabular}{c|c c|c c|c c||c c|c c|c c||c c|c c|c c}
\toprule
& \multicolumn{6}{c||}{Nanchang} & \multicolumn{6}{c||}{Manhattan}& \multicolumn{6}{c}{Hangzhou}\\
\cline{2-7}
\cline{8-13}
\cline{14-19}
& \multicolumn{2}{c}{Horizon 1} & \multicolumn{2}{c}{Horizon 3} & \multicolumn{2}{c||}{Horizon 5}& \multicolumn{2}{c}{Horizon 1} & \multicolumn{2}{c}{Horizon 3} & \multicolumn{2}{c||}{Horizon 5}& \multicolumn{2}{c}{Horizon 1} & \multicolumn{2}{c}{Horizon 3} & \multicolumn{2}{c}{Horizon 5}\\
\cline{2-7}
\cline{8-13}
\cline{14-19}
 &RMSE & MAPE & RMSE & MAPE & RMSE  & MAPE& RMSE & MAPE & RMSE & MAPE & RMSE  & MAPE & RMSE & MAPE & RMSE & MAPE & RMSE  & MAPE\\ 

\midrule
\midrule
HA &4.91 &23.18 &7.79 &25.73 &10.41 &29.97 &3.86&13.80 &5.45 &15.25 &6.76 &17.96& 4.72&21.41 &7.47 &23.36 &10.06 &27.21\\
ARIMA & 4.42&24.30 &6.90 &26.00 &9.58 &30.77& 3.78&14.02 &5.26 &15.23 &6.47 &17.95&4.36 &22.76 &6.62 &23.95 &9.07 &27.75 \\
\midrule
LSTM& 3.71 &17.17 &5.45 &22.40 &7.20 &28.10& 3.68 & 12.30 &5.04 &16.81 &5.41 &18.60& 3.55&16.52& 5.07 & 22.00& 6.79& 27.83\\
 LR&3.19 &18.78 &5.02 &24.70 &6.80 &30.07 & 2.68&12.93 &4.43 &18.56 &5.70 &22.61& 3.24&19.31 &4.91 &25.45 &6.73 &32.40\\
 XGBoost& 2.85& 14.74&4.91 & 20.70 & 6.86 &26.23 & 2.53&9.87 &4.31 &15.17 &5.45 &19.33& 2.94& 14.92& 4.84&20.95 &6.79 &27.48\\
\midrule
 DCRNN& 3.98& 18.91 &5.48 &26.42 &7.38 &31.94& 3.92& 14.31 & 5.42& 21.48 &7.01 & 29.56& 3.85&20.19 &5.44 &27.27 &7.49 &35.14 \\
 STGCN& 11.35& 33.35 &12.59 &37.18 &13.71 &38.90& 8.51& 25.40 & 9.29 &27.85 & 9.94 & 30.03&10.78 &33.30 &12.00 &36.52 &13.22 &39.94\\
 STDEN & 15.05 & 37.43  & 16.11  & 39.18  & 17.34  & 40.37  &  7.38  & 25.97   &  12.26    &  35.32  &  15.79   &  43.21 & 9.06  & 28.92  & 10.14  & 32.14  & 11.78  & 35.01\\
\midrule

 FDTI& \textbf{1.30} &\textbf{6.55} &\textbf{4.17} &\textbf{19.18}&\textbf{6.50} &\textbf{25.34}&\textbf{1.20} &\textbf{4.84} & \textbf{3.62} &\textbf{13.63} &\textbf{5.22} &\textbf{17.75}&\textbf{1.46} &\textbf{7.20} &\textbf{4.40} &\textbf{20.44} &\textbf{6.65} &\textbf{25.95}\\

\bottomrule
\bottomrule
\end{tabular}
}
\end{table*}

\begin{figure}[!tp]
    \centering
    \includegraphics[width=0.99\linewidth]{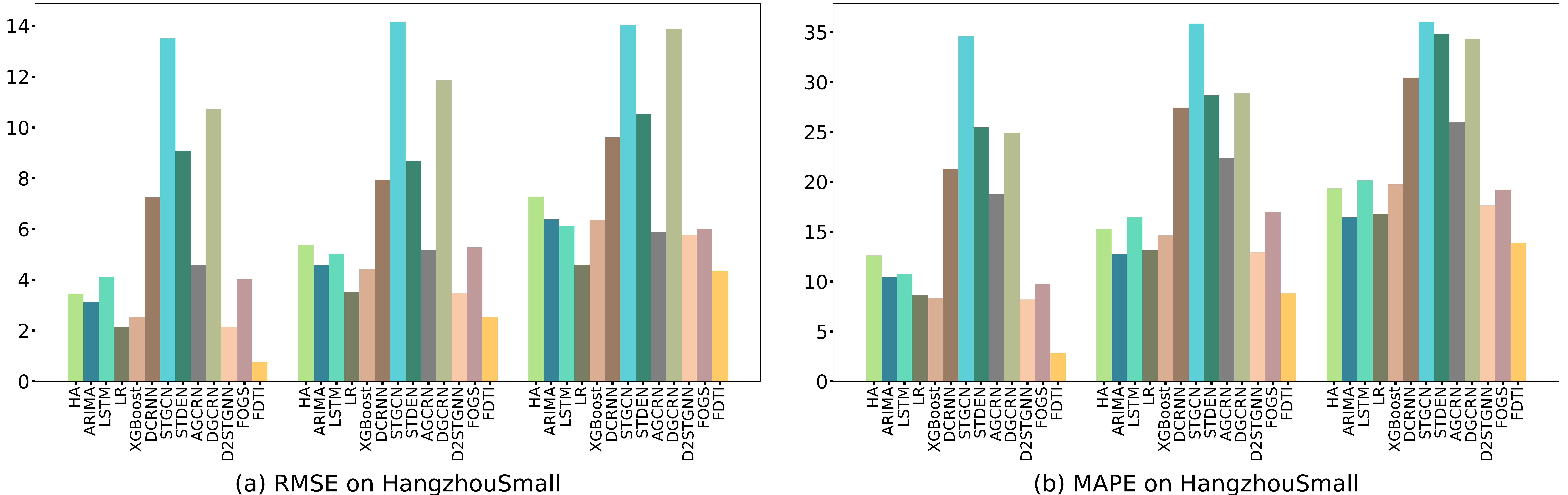}
    \caption{The performance of different methods w.r.t. RMSE and MAPE on Hangzhou-Small under horizon 1, 3 and 5. Horizon means the number of forecasting steps and one horizon means one minute. The lower the RMSE and MAPE are, the better. FDTI achieves the best performance.}
    \label{fig:hzsmall}
\end{figure}

The results of the comparison between FDTI and baselines are shown in Table~\ref{tab:performance} and Fig~\ref{fig:hzsmall}, where Table~\ref{tab:performance} shows the performance on three large-scale traffic datasets and Fig~\ref{fig:hzsmall} shows the performance on a small traffic dataset. 
On all four datasets with different scales, our proposed model outperforms all baseline methods in both single-step inference and multi-step inference.
The good performance indicates that the dynamic correlation modeling and the conservative traffic state inference based on flow-in and flow-out volume help the model grasp the intrinsic pattern of traffic. 
Note that other deep learning baselines perform worse than the traditional statistic methods and regression-based methods. 
This indicates that the dynamic correlation between traffic nodes could not be captured by simply stacking convolutional, recurrent, or adaptive mechanisms.
Another reason for the bad performance is that these GNN-based methods tend to yield smooth predictions on the nonsmooth dataset.
We will discuss the smoothness in detail later.
Besides, all of the adaptive methods are not able to run on large-scale datasets for their huge cost.
Hence, they are only evaluated on Hangzhou-Small.

\subsection{Graph Smooth Analysis}

In this part, we explain the reason why previous GCN-based methods yield unsatisfying results by analyzing the smoothness of datasets and prediction results.

\begin{figure}[!tb]
    \centering
    \includegraphics[width=.85\linewidth]{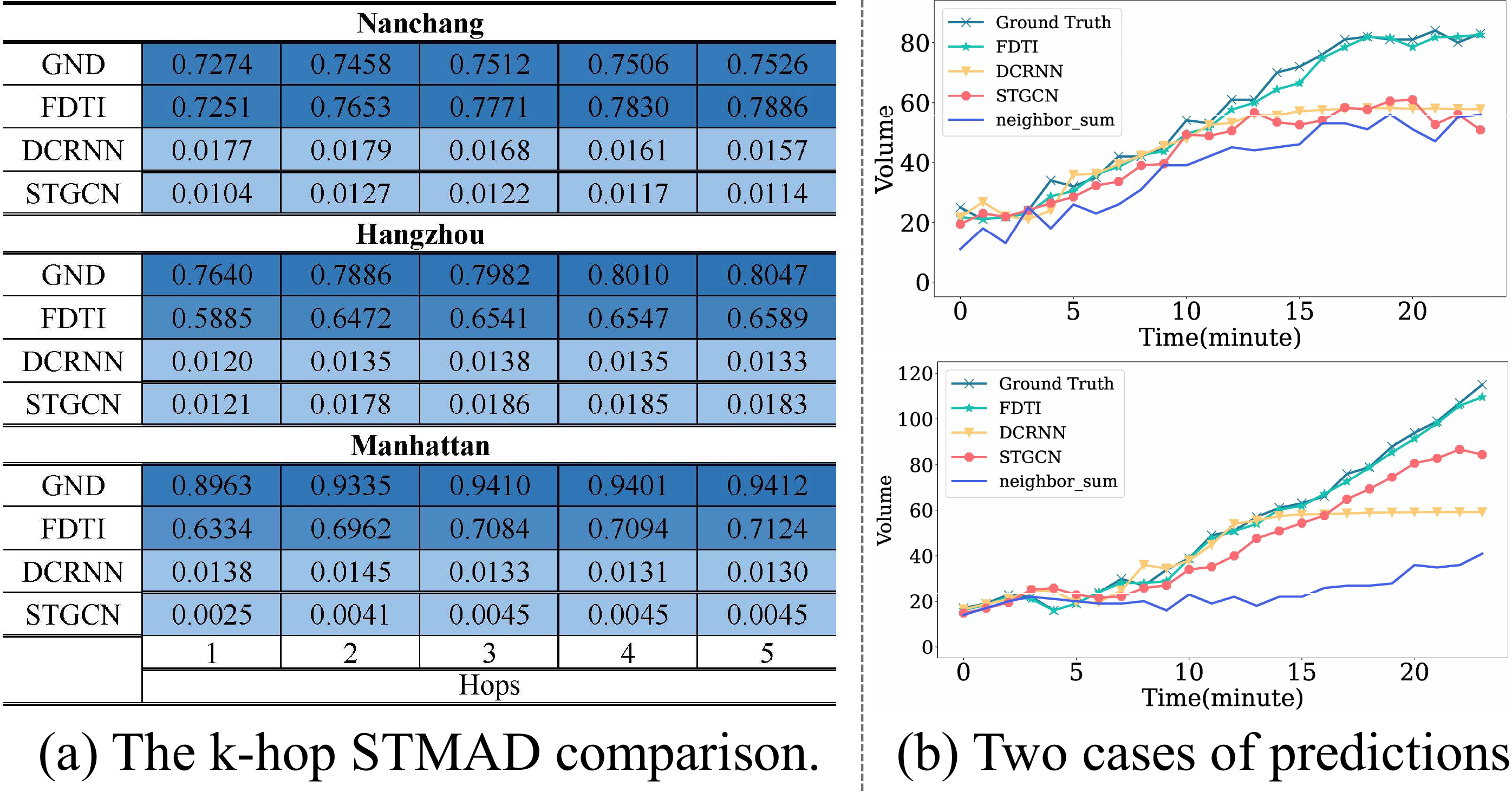}
    \caption{(a) The k-hop \textit{STMAD} comparison between GND (Ground Truth) and the predictions of several methods. (b) Cases of the predicted volume of several methods and ground truth.}
    \label{fig:RQ33}
\end{figure}

To quantitatively measure the smoothness over spatial temporal graphs, 
we leverage the \textit{STMAD} (Spatial-Temporal Mean Average Distance) based on MAD~\cite{chen2020measuring}. 
The MAD evaluates the smoothness of a given static graph with node features, and lower MAD indicates the graph is smoother.
Formally, given a spatial temporal graph in which each node contains a long time series data with length $\mathcal{T}$, we cut the time series data over a sliding window with length $\mathcal{P}$.
After that, the spatial temporal graph is cut into $\mathcal{\frac{T}{P}}$ subgraphs. 
The feature $H$ of each subgraph is the aligned partial time series data with length $\mathcal{P}$, i.e., $H\in\mathbb{R}^{N\times \mathcal{P}}$.
Then, we define the $k$-hop \textit{STMAD} as follows.
\begin{equation}
    STMAD^k = \frac{1}{\mathcal{\frac{T}{P}}}\sum_{m=1}^{\mathcal{\frac{T}{P}}}MAD^k_m\\
\end{equation}
\begin{equation}
    MAD^k_m = \frac{1}{N}\sum_{i=1}^N\frac{\sum_{j\in \mathcal{N}_k(i)}{(1-\frac{H_m^i\cdot H_m^j}{|H_m^i|\cdot |H_m^j|}})}{{|\mathcal{N}_k(i)|}}
\end{equation}
Here $STMAD^k$ means $k$-hop \textit{STMAD} and it is the average of the $k$-hop \textit{MAD} of all subgraphs.
$MAD^k_m$ is the $k$-hop \textit{MAD} of $m$-th subgraph and it is essentially the average cosine distance between nodes and their $k$-hop neighbors.
The $k$-hop neighbors set of node $i$ of $m$-th subgraph and its corresponding time series are denoted as $\mathcal{N}_k(i)$ and $H^i_m$ respectively.

We show the comparison of \textit{STMAD} between the ground truth data and the prediction result yielded by several methods in Fig~\ref{fig:RQ33}(a).
Among all the three datasets, We can observe that the \textit{STMAD} of ground truth is large due to its non-smoothness.
The non-smoothness could also be observed in the prediction result of FDTI, indicating that FDTI preserves the original traffic pattern of the ground truth data, thus making accurate predictions. 
On the contrary, the \textit{STMAD} of STGCN and DCRNN is much smaller than the \textit{STMAD} of ground truth data.
Furthermore, the \textit{STMAD} of STGCN and DCRNN is similar to the previous smooth datasets (METR-LA, PEMS-BAY, PEMSD7) as shown in Fig~\ref{fig:intro}(b).
This result explains that previous methodologies such as STGCN and DCRNN could yield satisfying results on the previous smooth datasets since these methodologies have a high tendency to make smooth predictions despite the smoothness of the input data. However, when it comes to unsmooth datasets, they make predictions with high errors.

Two examples of the smoothness of STGCN and DCRNN are shown in Fig~\ref{fig:RQ33}(b).
It shows the ground truth and prediction volume of a node along with the sum volume of its neighbors.
We could observe that STGCN and DCRNN make relatively reasonable predictions at the beginning since the ground truth volume is similar to the sum volume of neighbors.
As the traffic flow goes on, the gap between Ground Truth and the sum volume of neighbors increases, while STGCN and DCRNN fail to follow the Ground Truth.
Being consistent with the fact that smoothing is the essential nature of the GCN design~\cite{chen2020measuring}, this phenomenon indicates that STGCN and DCRNN tend to average the volume of a node and its neighbors and use the result as the prediction.
As a result, the prediction is smooth and a big error exists.
On the contrary, FDTI keeps consistent with the ground truth value, which is similar to the STMAD comparison.

To sum up, these two comparisons show FDTI performs admirably in the non-smooth situation.
We owe this excellent property to the conservative traffic transitions as shown in Eq~\eqref{eq:multilayer}.
This equation shows that FDTI predicts the first-order derivative of the ground truth and is hence resistant to oversmoothness.

\subsection{Ablation Study}

\begin{figure}[!t]
    \centering
    \includegraphics[width=0.9\linewidth]{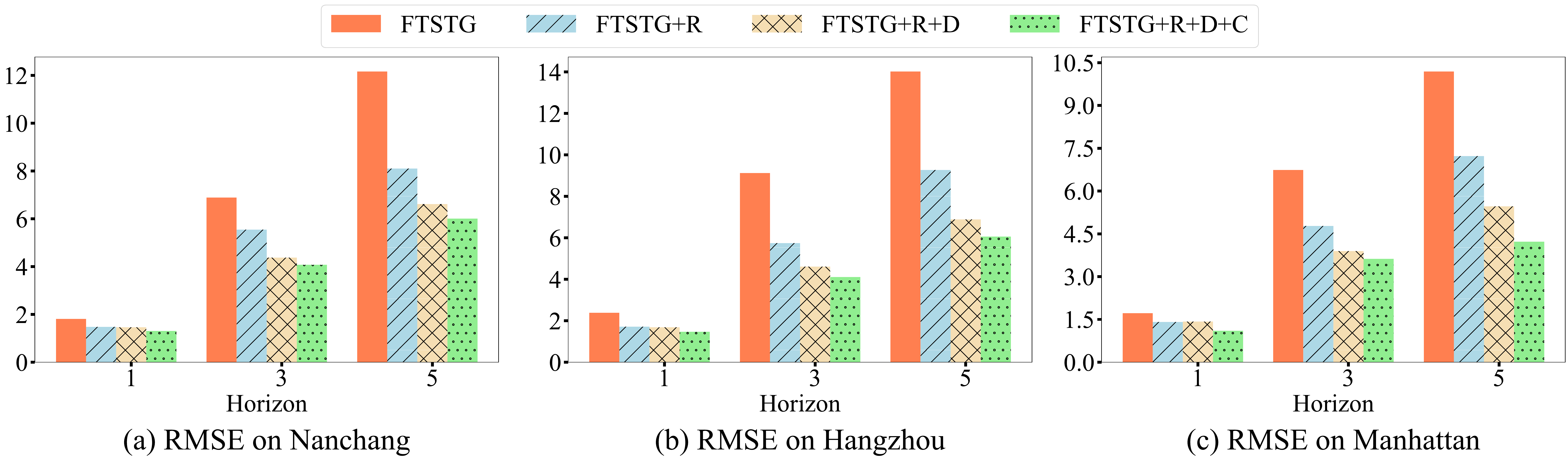}
    \caption{Performance on the multi-step inference of different variants of FDTI on three datasets. 
    Horizon means the number of forecasting steps.}
    \label{fig:ablation}
\end{figure}

For FDTI, there are four main designs including FTSTG that models the traffic dynamics, roadnet-enriched features (denoted as R) that help model capture traffic dynamics, the discount mechanism (denoted as D) that reduces the accumulated error of volume inference, and dynamic mobility convolution (denoted as C) that simulates the flow of vehicles. To validate these components, we design four variants by adding blocks sequentially: FTSTG, FTSTG+R, FTSTG+R+D, and FTSTG+R+D+C. Specifically, FTSTG+R+D+C equals FDTI because it has all of these four components. 

Results are shown in Fig~\ref{fig:ablation} from which we could observe that adding each module can induce further improvement. 
The improvement induced by adding the roadnet-enriched features (R) is due to
that adding traffic-dynamic-related features 
helps the model aggregate richer information. 
The performance of multi-step inference is improved by adding the discount mechanism (D), which indicates that the cumulative error could not be neglected and the discount mechanism tackles this error well. 
Adding dynamic mobility convolution (C) also brings notable performance gain. 
This demonstrates that considering the dynamic edges contributes to the fine-grained traffic dynamics between nodes.

\subsection{Scalability}

In this part, we explore the scalability of datasets and models. 
Then, we explain why previous adaptive methods fail to run on our datasets.

\paragraph{City-scale Datasets and Experiments}
To the best of our knowledge, we are the first to complete the city-level traffic state inference. These three city-level traffic datasets Nanchang, Hangzhou, and Manhattan cover more than 2,000 intersections and 18,000 nodes as shown in Table~\ref{tab:datasets}. In comparison, we have summarized the datasets used in previous literature as in Table~\ref{tab:scale_data}. It is easy to observe that our datasets are at least 10 times larger than previously wildly-used datasets in terms of the number of nodes.

\paragraph{Model Scalability}

\begin{figure}[!t]
    \centering
    \includegraphics[width=0.85\linewidth]{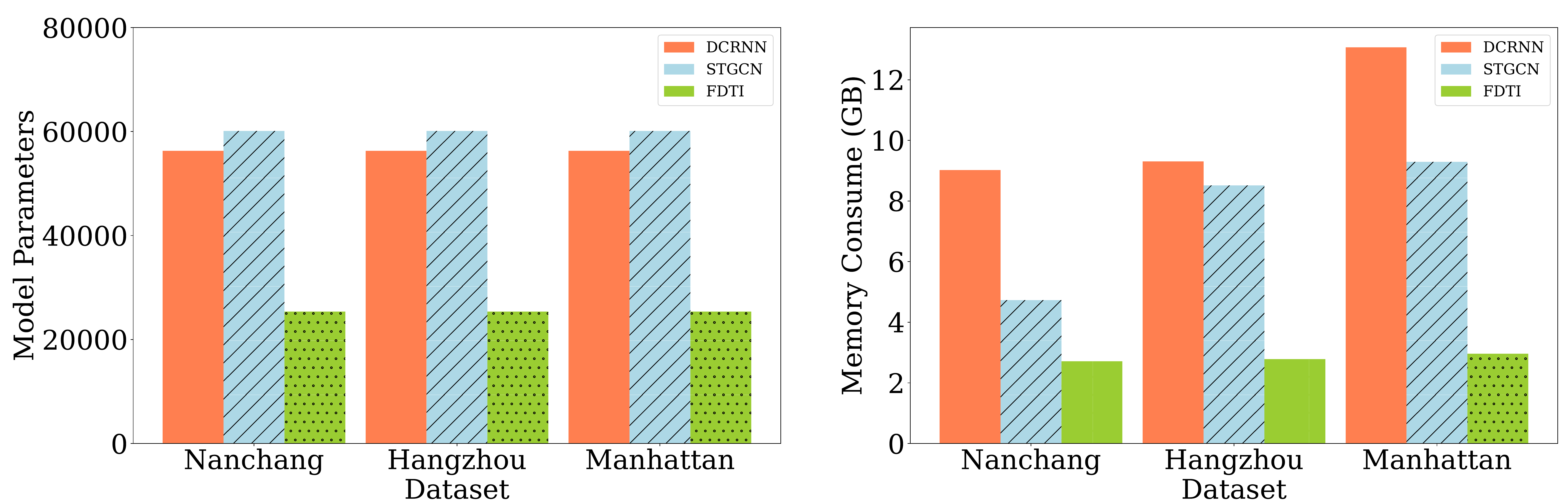}
    \caption{Model parameter size and memory cost.}
    \label{fig:scale}
\end{figure}

\begin{table}[t]
\caption{The scale of datasets. The upper 8 datasets are wildly used by previous research. The lower 4 datasets downside are the datasets of this paper.}
\label{tab:scale_data}
\setlength\tabcolsep{2pt}
\centering
\resizebox{0.6\linewidth}{!}{
\begin{tabular}{c c c c}
\toprule
Dataset     &  \# of nodes & Dataset & \# of nodes\\ 
\midrule
\midrule
PeMSD7(M)~\cite{yu2017spatio}     &  228 & METR-LA~\cite{li2017diffusion} & 207\\
PeMSD7(L)~\cite{yu2017spatio} & 1,026 & PEMS-BAY~\cite{li2017diffusion} & 325\\
PeMSD4~\cite{guo2019attention} & 307 & Xiamen~\cite{zheng2020gman} & 95\\
PeMSD8~\cite{guo2019attention} & 170 & PeMS3~\cite{song2020spatial} & 358\\
\midrule
\midrule
Nanchang & 18,072 & Manhattan & 40,026\\
Hangzhou & 32,772 & Hangzhou-Small & 1,944\\
\bottomrule
\end{tabular}
}
\end{table}

The space complexity of FDTI is $O(d\times d)$ and thus the model size is independent of the input graph scale. 
Furthermore, FDTI is an inductive graph learning method due to the special construction of FTSTG that limited hops of neighbors are required for predicting future traffic.
Benefiting from this, our model could deal with large-scale traffic graphs with decent parameter efficiency. 

Most of the previous deep spatial temporal methods~\cite{guo2021hierarchical,oreshkin2021fc,lee2021learning,shao2022decoupled,rao2022fogs} based on adaptive mechanism fail to run on large-scale datasets. 
Firstly, they have at least $O(N\times d)$ parameters as node embeddings, which is not parameter-efficient.
Secondly, the space complexity of calculating the similarity matrix of these methods is $O(N^2)$, which is unacceptable for a large graph. 
To validate their performance in the fine-grained setting, we extract Hangzhou-Small dataset and implement some of them as baselines. 

For the rest deep learning based methods, we select DCRNN (best performance), STGCN (most efficient), and FDTI (this paper) and compare the model efficiency on the large-scale datasets with the same number of hidden states and layers as shown in Figure~\ref{fig:scale}. We can observe that FDTI has a similar number of parameters while FDTI consumes much less memory.

\section{Conclusion}
In this paper, we have worked on a brand-new fine-grained traffic volume prediction problem. 
We demonstrate that the traffic signal significantly influences the correlation between neighboring roads.
To address the problems, We propose a novel method FDTI that models the influence of traffic signal and capture the fine-grained traffic dynamics.
Extensive experiments are conducted on large-scale traffic datasets, where FDTI outperforms other baselines and shows good properties such as resistance to oversmoothness. 
We believe that FDTI can better support real-world downstream applications such as traffic policy making.

\section*{Acknowledgement}

This work was sponsored by National Key Research and Development Program of China under Grant No.2022YFB3904204, National Natural Science Foundation of China under Grant No. 62102246, 62272301, and Provincial Key Research and Development Program of Zhejiang under Grant No. 2021C01034.

\bibliographystyle{splncs04}
\bibliography{fdti}
\clearpage
\end{document}